\documentclass[letterpaper, 10 pt, conference]{ieeeconf}  

\IEEEoverridecommandlockouts                              

\title{\LARGE \bf
PlaneHEC: Efficient Hand-Eye Calibration for Multi-view Robotic Arm via Any Point Cloud Plane Detection
}

\author{Ye Wang$^{1}$, Haodong Jing$^{1}$, Yang Liao$^{1}$, Yongqiang Ma$^{1}$, Nanning Zheng$^{1\dagger}$
\vspace{-0.5em}
\thanks{$\dagger$ Corresponding author.}
\thanks{$^{1}$ Y. Wang, H. Jing, Y. Liao, Y. Ma, and N. Zheng are with National Key Laboratory of Human-Machine Hybrid Augmented Intelligence, Institute of Artificial Intelligence and Robotics, Xi'an Jiaotong University. 
        {\tt\small \{a1445676266, jinghd, bluemagpie\}@stu.xjtu.edu.cn, musayq@xjtu.edu.cn, nnzheng@mail.xjtu.edu.cn}}%
}

\usepackage{multirow}
\usepackage{booktabs} 
\usepackage{graphicx}

\usepackage{amsmath}

\makeatletter
\renewcommand{\maketag@@@}[1]{\hbox{\m@th\normalsize\normalfont#1}}%
\makeatother

\begin{document}

\maketitle
\thispagestyle{empty}
\pagestyle{empty}

\begin{abstract}
Hand-eye calibration is an important task in vision-guided robotic systems and is crucial for determining the transformation matrix between the camera coordinate system and the robot end-effector. Existing methods, for multi-view robotic systems, usually rely on accurate geometric models or manual assistance, generalize poorly, and can be very complicated and inefficient. Therefore, in this study, we propose PlaneHEC, a generalized hand-eye calibration method that does not require complex models and can be accomplished using only depth cameras, which achieves the optimal and fastest calibration results using arbitrary planar surfaces like walls and tables. PlaneHEC introduces hand-eye calibration equations based on planar constraints, which makes it strongly interpretable and generalizable. PlaneHEC also uses a comprehensive solution that starts with a closed-form solution and improves it with iterative optimization, which greatly improves accuracy. We comprehensively evaluated the performance of PlaneHEC in both simulated and real-world environments and compared the results with other point-cloud-based calibration methods, proving its superiority. Our approach achieves universal and fast calibration with an innovative design of computational models, providing a strong contribution to the development of multi-agent systems and embodied intelligence.


\end{abstract}

\section{Introduction}

Hand-eye calibration is a key robot vision technique in robotic systems \cite{enebuse2021comparative}, which solves the relative position and attitude between the camera and the robotic arm \cite{jiang2022overview}. Hand-eye calibration technology is widely used in industrial robots, service robots and many other applications. The significance of hand-eye calibration not only lies in improving the accuracy and reliability of robot operation, but also lays the foundation for the intelligent development of robot systems. By combining deep learning algorithms, hand-eye calibration technology can continuously improve the adaptability and autonomy of robots in complex environments. Hand-eye calibration has a wide application prospect in autonomous driving, human-robot collaboration and other cutting-edge fields, which strongly promotes the robot system towards the development of embodied intelligence \cite{gupta2021embodied}.

In recent years, hand-eye calibration methods can be broadly classified into two categories: geometric model-based methods \cite{zhi2024unifying,liu2024gbec} and optimization solution-based methods \cite{kaiser2008extrinsic,shah2012overview, su2022research}. Geometric modeling methods solve for the transformation matrix between the robot arm and camera by obtaining multiple known poses. Common geometric methods include Tsai-Lenz method \cite{tsai1985efficient} and Horaud method \cite{horaud1995hand}, which can provide more accurate calibration results. However, as the complexity of the scene increases, these methods show shortcomings in terms of sensor errors as well as the accuracy required for robot motion. The hand-eye calibration method based on 3D vision sensors proposed by Fu et al. \cite{fu2020hand} improves the accuracy of pose estimation but still relies on the assumptions of accurate sensor data and geometric models. The optimization-based solution methods minimize the pose error between the end-effector and the camera through iterative optimization, the Schur matrix method \cite{liu2019robust} proposes hand-eye calibration methods with enhanced robustness to maintain higher accuracy against noises via Schur matrices. However, such methods tend to rely on the quality of the initial solution and are susceptible to local optimal solutions \cite{fu2022hand}. Using depth cameras for accuracy assessment \cite{xing2022reconstruction,xing2023reghec,liu2020fast}, despite some progress, complex data such as point clouds still pose computational convergence problems.

\begin{figure}
        \centering
        \includegraphics[width=0.9\linewidth]{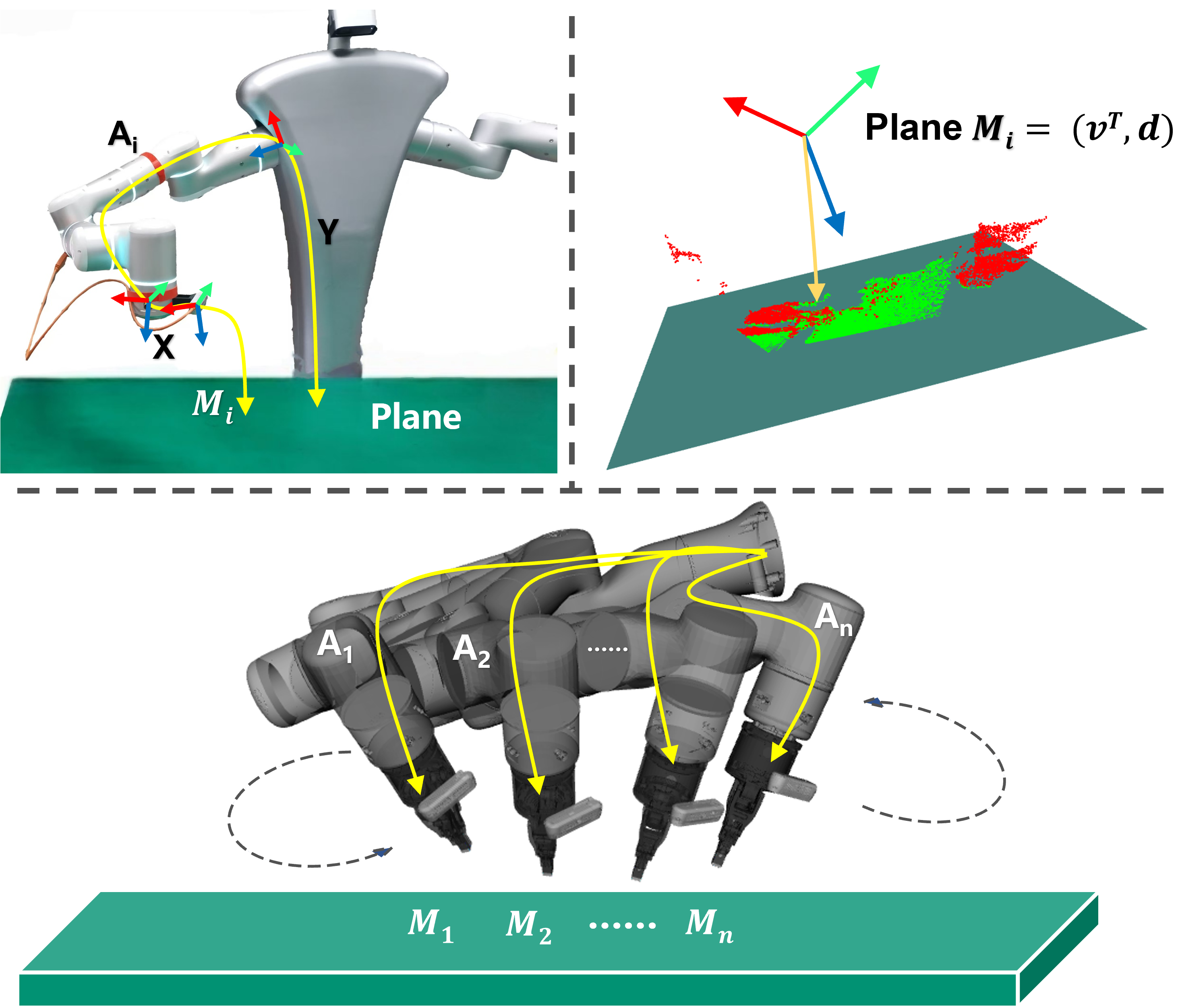}
        \vspace{-0.5em}
        \caption{\textbf{Overview of PlaneHEC.} The depth camera fixed on the robot arm observes the planar surface from the multi-view point cloud. Then the plane detection algorithm computes the plane's equation, which is utilized to estimate the transformation between the camera and the robotic arm.
        }
        \vspace{-1.2em}
        \label{fig:overview}
\end{figure}

To overcome the limitations of existing methods, this paper proposes PlaneHEC, a hand-eye calibration approach based on point cloud plane detection as shown in Figure~\ref{fig:overview}. We argue that using planes as calibration objects has several advantages \cite{carlson2015six} and is particularly well-suited for real-world applications. First, PlaneHEC only requires the plane to be relatively flat, without the need for a high-precision standard plane. Such planes are common in real-world environments, like walls or tabletops, and all fit our approach. PlaneHEC eliminates the need for additional high-precision calibration models, making it much more generalized. Second, the plane detection approach based on point clouds is highly generalizable, as it does not require 
a pre-trained model but merely sets the camera's field of view, allowing it to adapt to various environments. Third, PlaneHEC takes all the points into account to find the best plane, making the best use of a depth camera. A plane is derived from a large set of observed points, which helps mitigate the impact of camera depth noise on the results. Additionally, the solution for the plane can be obtained through a closed-form solution, avoiding the dependence on initial values typical in ICP algorithms \cite{Segal2009GeneralizedICP}, resulting in a more robust and stable performance.

The contributions of this paper are as follows:
\begin{itemize}
\item 
We propose PlaneHEC, a plane-detection-based hand-eye calibration method tailored for 3D point cloud cameras, and reformulate the hand-eye calibration problem for 3D sensors from the perspective of plane detection. 
\item 
We introduce a complete solution to the new calibration equations, which first obtains an initial value through a closed-form solution and then refines the solution using 
the iterative Lie algebra optimization method.
\item 
We validate the robustness and accuracy of PlaneHEC through experiments using a low-cost structured light camera (RealSense-D435) on the FLEXIV-Rizon 4 platform. The sufficient experiments proved PlaneHEC to be the state-of-the-art hand-eye calibration method in terms of accuracy and runtime. \end{itemize}

\section{Related Work}
Hand-eye calibration has been extensively studied since the 1980s \cite{shiu1987calibration, tsai1989new}, beginning with the well-known kinematic loop in the form of $AX=XB$ first proposed by Shiu and Ahmad \cite{shiu1987calibration}. This formulation is universal, as it describes rigid body motion, making it independent of the sensor type or its operating principles~\cite{xing2022reconstruction}.
Most traditional methods rely on 2D image data and techniques like planar calibration patterns to estimate camera motion \cite{jiang2022overview, liu2024gbec, antonello2017fully, jin2023simultaneous, jin2023online}.  However, they are not well-suited for depth cameras, particularly those lacking RGB outputs \cite{xing2022reconstruction}. With the advent of 3D sensors, point cloud-based hand-eye calibration methods are crucial.

Recently, point cloud-based calibration methods have emerged, such as those utilizing high-precision 3D models \cite{fu2020hand, fu2022hand, liu2020fast, yang2018robotic, wu2023simple, kahn2014hand, xie2021general}. 
However, these approaches require expensive calibration devices and are unsuitable for scenarios where a model cannot be obtained. Other methods based on ICP (Iterative Closest Point) \cite{Segal2009GeneralizedICP} reconstruction errors have also been explored \cite{xing2022reconstruction, li2024automatic, peters2024robot}, but ICP methods suffer from poor convergence, dependence on initial values, and sensitivity to noise, making them less effective for low-cost cameras. There are also calibration methods based on matching 3D models of the robot arm \cite{li2024automatic, sefercik2023learning}. However, these methods depend on the camera being able to observe the robot arm, making them impractical in situations where the arm is covered by materials like fabric. Additionally, some methods rely on neural networks \cite{sefercik2023learning, chen2023easyhec}, which require retraining when applied to different robot models, limiting their generalizability.

In previous uses of planar constraints, Kaiser \cite{kaiser2008extrinsic} manually measured the plane to calibration, but this is prone to low accuracy and requires additional manual effort. Carlson \cite{carlson2015six} used the plane constraint approach for line laser scanners. Applying this method to 3D cameras poses challenges, like failing to remove outliers and inconsistent frame weights from varying point distributions. 
Planar features are also used in other fields, such as SLAM and 3D reconstruction. For example, Yanyan Li~\cite{li2021rgb} used point, line, and surface features to improve the accuracy of building maps in structured scenes. 
Therefore, we have improved the 3D camera-based planar constraint method and introduced a new calibration equation in the form $Y=MXA$. Unlike previous equations \cite{tsai1989new}, our $Y$ and $M$ are 1×4 vectors, requiring new solution approaches. To solve this, we referenced methods \cite{li2010simultaneous, shah2013solving} involving the Kronecker product, establishing a least-squares error formulation and deriving a closed-form solution for our new equation. Additionally, we further reduced errors by employing Lie Theory \cite{park1994robot} for differentiation and iteratively optimizing using the Newton-Gauss method, thereby improving solution accuracy.

\section{Methodology}

\begin{figure*}
        \centering
        \includegraphics[width=0.91\linewidth]{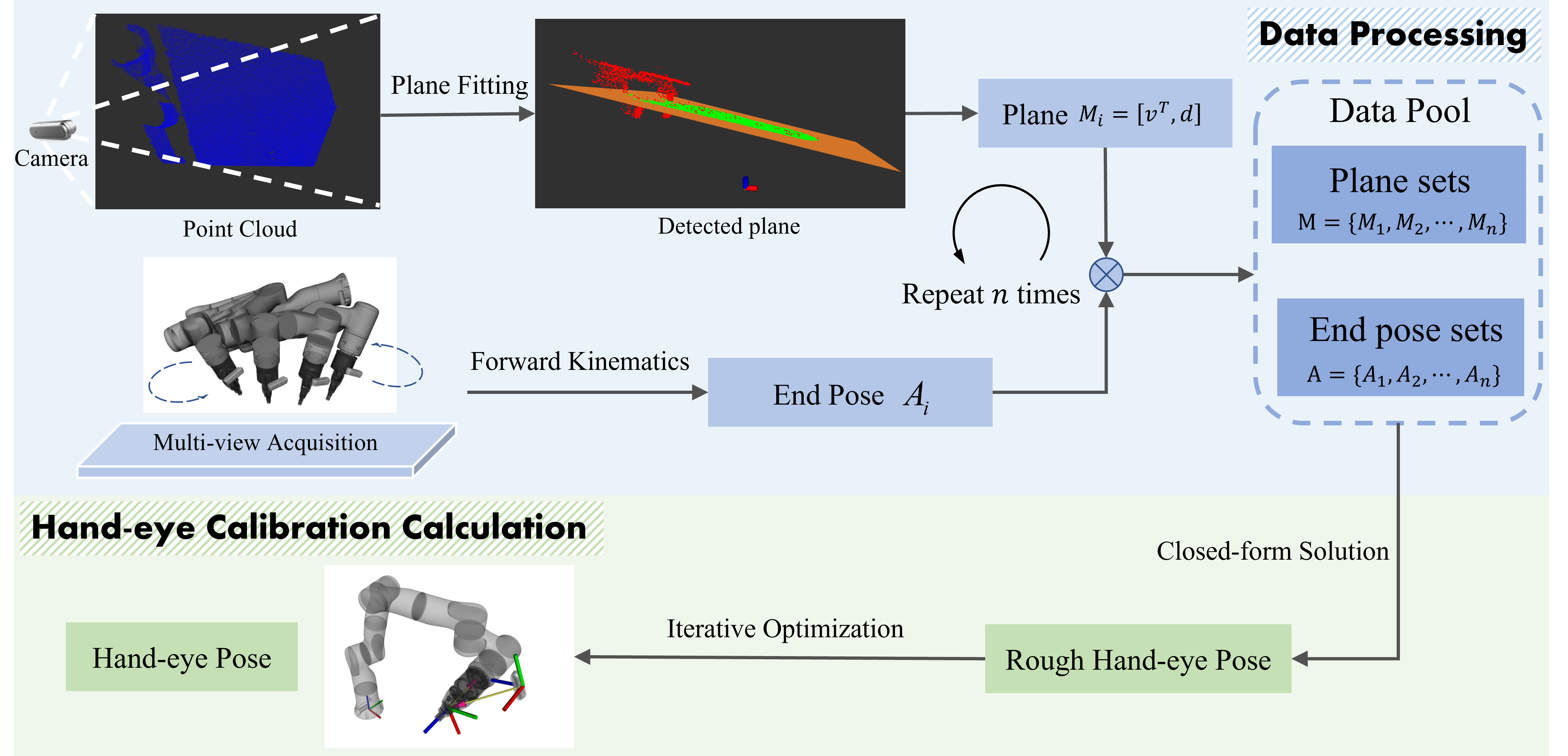}
        \vspace{-0.8em}
        \caption{
                \textbf{The framework of the proposed PlaneHEC method.} The robotic arm observes the planar surface from multiple viewpoints, while the plane detection algorithm calculates the plane's normal vector and distance. To solve the plane constraints, we employ a two-step approach, starting with a closed-form solution, followed by iterative refinement, which ultimately estimates the relative pose between the camera and the robotic arm. 
        }
        \vspace{-1em}
        \label{fig:pipeline}
\end{figure*}

The framework of PlaneHEC, as shown in Figure~\ref{fig:pipeline}, is primarily composed of three key components:
\begin{itemize}
    \item \textbf{Plane Detection.} We collect the point cloud of a fixed plane from multiple views with the help of a RGB-D camera. Then, we adopt RANSAC \cite{martinez2022ransac} algorithm to derive the plane equation.
    \item \textbf{Closed-form Solution.} We calculate the least square solution, also known as the closed-form solution by solving linear equations derived from plane detection.
    \item \textbf{Iterative Optimization.} We optimize the closed-form solution iteratively by means of Gauss–Newton method and Lie group.
\end{itemize}

\subsection{Issue Definition of PlaneHEC} 
The equation of a plane in space can be expressed as
\vspace{-0.5em}
\begin{equation}
ax+by+cz+d=0,\vspace{-0.5em}
\end{equation}

whose normal vector is $\mathbf{v}=(a,b,c)^{T}$, following the constraint $||\mathbf{v}||=1$ to ensure it is a unit vector. Let $M=(\mathbf{v}^{T},d)$, for any point on the plane, its homogeneous coordinates $\mathbf{p}=(x,y,z,1)^{T}$ should satisfy the equation
\vspace{-0.5em}
\begin{equation}
M\mathbf{p}=0.\vspace{-0.5em}
\end{equation}

Assume the plane is static to the base of the robot, put \emph{Eq.2} in base frame and we have
\vspace{-0.5em}
\begin{equation}
M_{base}\mathbf{p}_{base}=0,\vspace{-0.5em}
\end{equation}

where $M_{base}$ is invariant. Simple it seems, we can not directly obtain the homogeneous coordinates of points on the plane. When observing the plane from camera, we get
\vspace{-0.5em}
\begin{equation}
M\mathbf{p}_{camera}=0,
\end{equation}

where $p$ is put in camera frame. To transform the coordinates from camera frame into base frame, we have
\vspace{-0.5em}
\begin{equation}
MXA\mathbf{p}_{base}=0,\vspace{-0.5em}
\end{equation}

where $X$ and $A$ are transformation matrices from TCP frame to camera frame and from base frame to TCP frame. Let $Y=MXA$, every plane observed equation is equivalent with \emph{Eq.3}, which means the row vectors $Y$ and $M_{base}$ are linear. With the constraint that $||\mathbf{v}||=1$, $Y$ is either $M_{base}$ or its negative. By determining the direction of the normal vector, we can ensure that $Y$ always equals $M_{base}$. The matrix $A$ can be read from the robot, matrix $X$ is the target of hand-eye calibration. 
In order to improve robustness, define 
\vspace{-0.5em}
\begin{equation}
A_{average} = 
\begin{bmatrix}
I & \mathbf{t}_{average} \\
\mathbf{0}^{T} & 1 \\
\end{bmatrix}^{-1},\vspace{-0.5em}
\end{equation}
where $\mathbf{t}_{average}$ is the average of translation vectors in $A$. We multiply $A$ on the right by $A_{average}$ to make the translation vectors normalized. This won't hurt the nature of $A$, so we keep the name as it is. By observing the same plane from $n$ perspectives, we can obtain $n$ equations
\vspace{-0.5em}
\begin{equation}
Y=M_{i}XA_{i},
\end{equation}

where $i=1,2,...,n$. The PlaneHEC problem is described in \emph{Eq.7}. In this context, $Y$ and $M$ represent the plane, with dimensions of 1x4, while $X$ and $A$ represent the transformation matrices, with dimensions of 4x4. $X$ and $Y$ are the fixed unknowns to be solved, whereas $M_{i}$ and $A_{i}$ are the data obtained from multi-views, representing the plane and the robot arm poses. Our objective is to solve for $X$.
\vspace{-0.5em}
\subsection{Plane Detection}

We use RANSAC \cite{martinez2022ransac} algorithm to detect the plane and calculate its equation. To be specific, we randomly sample 3 points from the point cloud, and calculate their coplanar equation. Then, we calculate the distance of any other points to the plane and decide the number of inliers. If the number exceeds the threshold we set, the equation is adopted. Otherwise, repeat until one equation fits the condition.
\vspace{-0.5em}
\subsection{Closed-form Solution}
For \emph{Eq.7}, we can find the least square solution to $X$ by at least 4 equations, also known as the closed-form solution. Divide $X$ and $A_{i}$ into rotation matrices and translation vectors, \emph{Eq.7} can be rewritten as 
\vspace{-0.5em}
\begin{equation}\vspace{-0.5em}
\begin{aligned}
Y_{i}&=
\left[
\begin{array}{cc}
\mathbf{v}_{i}^{T} & d_{i}
\end{array}
\right]
\left[
\begin{array}{cc}
R_{X} & \mathbf{t}_{X} \\ 0 & 1
\end{array}
\right]
\left[
\begin{array}{cc}
R_{A,i} & \mathbf{t}_{A,i} \\ 0 & 1
\end{array}
\right]\\
&=
\left[
\begin{array}{cc}
\mathbf{v}_{i}^{T}R_{X}R_{A,i} & \mathbf{v}_{i}^{T}R_{X}\mathbf{t}_{A,i}+\mathbf{v}_{i}^{T}\mathbf{t}_{X}+d_{i}
\end{array}
\right].
\end{aligned}
\end{equation}
Thus, we can obtain $n-1$ equations
\begin{equation}\vspace{-0.5em}
\mathbf{v}_{j}^{T}R_{X}R_{A,j}=\mathbf{v}_{j+1}^{T}R_{X}R_{A,j+1},
\end{equation}
and 
\begin{equation}
\mathbf{v}_{j}^{T}R_{X}\mathbf{t}_{A,j}+\mathbf{v}_{j}^{T}\mathbf{t}_{X}+d_{j}=\mathbf{v}_{j+1}^{T}R_{X}\mathbf{t}_{A,j+1}+\mathbf{v}_{j+1}^{T}\mathbf{t}_{X}+d_{j+1},
\end{equation}
where $j=1,2,...,n-1$. We will start by solving \emph{Eq.9}. Let $vec(R_{x})$ be the vectorization form of $R_{X}$, \emph{Eq.9} can be rewritten as
\begin{equation}
(R_{A,j}^{T}\otimes\mathbf{v}_{j}^{T}-R_{A,j+1}^{T}\otimes\mathbf{v}_{j+1}^{T})vec(R_{x})=0,
\end{equation}
where $\otimes$ is the Kronecker product \cite{shah2013solving}. \emph{Eq.11} is a homogeneous system of linear equations, we need at least 3 equations to find the least square solution. Then, we applied SVD for orthogonalization and normalization, obtaining $R_{x}$. 
Once we find $R_{x}$, we can proceed to solve \emph{Eq.10}. Rewrite \emph{Eq.10} as 
\begin{equation}
(\mathbf{v}_{j}^{T}-\mathbf{v}_{j+1}^{T})\mathbf{t}_{X}=-\mathbf{v}_{j}^{T}R_{X}\mathbf{t}_{A,j}+\mathbf{v}_{j+1}^{T}R_{X}\mathbf{t}_{A,j+1}-d_{j}+d_{j+1}.
\end{equation}
\emph{Eq.12} is a system of non-homogeneous linear equations, whose least square solution is also easy to find. Now we have both $R_{x}$ and $t_{x}$, thus $X$ is found.
\vspace{-0.5em}
\subsection{Iterative Optimization}
\vspace{-0.5em}
In the process of finding the closed-form solution, we first deal with $R_{x}$ then $t_{x}$, which can lead to error propagation. Therefore, we use Gauss–Newton method to find $R_{x}$ and $t_{x}$ simultaneously. We define the objective function to be optimized as
\vspace{-0.5em}
\begin{equation}
\begin{aligned}
min\;G(X)=\sum_{j=1}^{n-1}||g_{j}(X)||^{2},\\
s.t.{\quad}X{\in}SE(3),
\end{aligned}
\end{equation}
where
\begin{equation}
g_{j}(X)=M_{j}XA_{j}-M_{j+1}XA_{j+1}.
\end{equation}
Divide $g_{j}$, $A_{j}$ and $A_{j+1}$ into 4 columns respectively, we have
\begin{equation}
g_{j}=\left[g_{j,1},g_{j,2},g_{j,3},g_{j,4}\right]
\end{equation}
and
\begin{equation}
A_{j}=
\left[
\begin{array}{cccc}
\mathbf{\alpha}_{j}&\mathbf{\beta}_{j}&\mathbf{\gamma}_{j}&\mathbf{\delta}_{j}\\
0&0&0&1
\end{array}
\right],
\end{equation}
where $\mathbf{\alpha}_{j}$, $\mathbf{\beta}_{j}$, $\mathbf{\gamma}_{j}$, $\mathbf{\delta}_{j}$ are 3$\times$1 vectors. $g_{j}$ can be divided into
\small{
\begin{equation}
\left\{
\begin{aligned}
g_{j,1} &= M_{j}X
\left(
\begin{bmatrix}
\mathbf{\alpha}_{i}\\
1
\end{bmatrix} -
\begin{bmatrix}
\mathbf{0} \\
1
\end{bmatrix}
\right)
-
M_{j+1}X
\left(
\begin{bmatrix}
\mathbf{\alpha}_{i+1} \\
1
\end{bmatrix} -
\begin{bmatrix}
\mathbf{0} \\
1
\end{bmatrix}
\right)\\
g_{j,2} &= M_{j}X
\left(
\begin{bmatrix}
\mathbf{\beta}_{j} \\
1
\end{bmatrix} -
\begin{bmatrix}
\mathbf{0} \\
1
\end{bmatrix}
\right)
-
M_{j+1}X
\left(
\begin{bmatrix}
\mathbf{\beta}_{j+1} \\
1
\end{bmatrix} -
\begin{bmatrix}
\mathbf{0} \\
1
\end{bmatrix}
\right)\\
g_{j,3} &= M_{j}X
\left(
\begin{bmatrix}
\mathbf{\gamma}_{j} \\
1
\end{bmatrix} -
\begin{bmatrix}
\mathbf{0} \\
1
\end{bmatrix}
\right)
-
M_{j+1}X
\left(
\begin{bmatrix}
\mathbf{\gamma}_{j+1} \\
1
\end{bmatrix} -
\begin{bmatrix}
\mathbf{0} \\
1
\end{bmatrix}
\right)\\
g_{j,4} &= M_{j}X
\begin{bmatrix}
\mathbf{\delta}_{j} \\
1
\end{bmatrix}
-
M_{j+1}X
\begin{bmatrix}
\mathbf{\delta}_{j+1} \\
1
\end{bmatrix}
\end{aligned}
\right.
.\end{equation}
}
\normalsize

$\mathbf{\alpha}_{j}$, $\mathbf{\beta}_{j}$, $\mathbf{\gamma}_{j}$ and $\mathbf{\delta}_{j}$ can all be seen as points in space, so it is easy to derive the Jacobian Matrix of $g_{j,1}$, $g_{j,2}$, $g_{j,3}$, $g_{j,4}$ respectively  with perturbation model
\vspace{-0.5em}
\small{
\begin{equation}
\left\{
\begin{aligned}
J_{j,1}(\xi,X)={\quad}&M_{j}(X\alpha_{j})^{\odot}&-&M_{j}(X\mathbf{0})^{\odot}\\-&M_{j+1}(X\alpha_{j+1})^{\odot}&+&M_{j+1}(X\mathbf{0})^{\odot}\\
J_{j,2}(\xi,X)={\quad}&M_{j}(X\beta_{j})^{\odot}&-&M_{j}(X\mathbf{0})^{\odot}\\-&M_{j+1}(X\beta_{j+1})^{\odot}&+&M_{j+1}(X\mathbf{0})^{\odot}\\
J_{j,3}(\xi,X)={\quad}&M_{j}(X\gamma_{j})^{\odot}&-&M_{j}(X\mathbf{0})^{\odot}\\-&M_{j+1}(X\gamma_{j+1})^{\odot}&+&M_{j+1}(X\mathbf{0})^{\odot}\\
J_{j,4}(\xi,X)={\quad}&M_{j}(X\delta_{j})^{\odot}&-&M_{j+1}(X\delta_{j+1})^{\odot}
\end{aligned}
\right.
,\end{equation}
}
\normalsize
where $\xi$ is the left perturbation on $X$, and $(\cdot)^{\odot}$ represents for the result of derivation. For $X{\in}SE(3)$ and 3D point $\mathbf{p}$, 

\begin{equation}
(X\mathbf{p})^{\odot} =
\begin{bmatrix}
    \textbf{I} & -(R\textbf{p} +\textbf{t})^{\land} \\
    \textbf{0}^\top & \textbf{0}^\top
\end{bmatrix}_{4\times6},
\end{equation}

where $(\cdot)^{\land}$ is the skew-symmetric matrix. Arrange all the $g$ and $J$ in matrices, we have
\begin{equation}
\mathbf{g}=
\left[
g_{1,1}, g_{1,2}, ..., g_{1,4}, g_{2,1}, ..., g_{n-1,4}
\right]
_{4(n-1)\times1}^{T},
\end{equation}
and
\begin{equation}
\mathbf{J}=
\left[
J_{1,1}, J_{1,2}, ..., J_{1,4}, J_{2,1}, ..., J_{n-1,4}
\right]
_{4(n-1)\times6}^{T}.
\end{equation}
According to Gauss–Newton method, the normal equation is 
\begin{equation}
\mathbf{J}^{T}\mathbf{J}{\Delta}X=-\mathbf{J}^{T}\mathbf{g},
\end{equation}
where ${\Delta}X$ is a small perturbation in the form of Lie algebra. We iterate over ${\Delta}X$ by calculating 
\begin{equation}
{\Delta}X = -(\mathbf{J}^{T}\mathbf{J})^{-1}\mathbf{J}^{T}\mathbf{g},
\end{equation}
until ${\Delta}X$ is small enough. The initial $X$ can have a big impact on the optimization results, so we choose the closed-form solution as the initial value. Our analysis reveals that inaccuracies in plane detection exert considerable influence on closed-form solutions, while iterative methodologies effectively mitigate such perturbations.

\section{Experiments}
\subsection{Experimental Setup}
\textbf{Implementation Details. }We implement our experiments on a desktop with i5-12500H CPU. We use FLEXIV-Rizon 4 robotic arm for real environment tests, whose repeated positioning accuracy can reach ±0.05mm. The RealSense-D435 camera with 1280×720 resolution is used in the experiment. This camera has a relative accuracy of less than 2\%, and we collected data with points' depth ranging from 0.3m to 0.8m, of which the accuracy is approximately 10mm~\cite{ahn2019analysis}. We only use the depth data for computation, but we employ both depth data and RGB data to better present the final experimental results.

\textbf{Evaluation Metrics. }We adopt $e_{r}$, $e_{t}$, $e_{xy}$ and $e_{z}$ to assess methods' accuracy. $e_{r}$ and $e_{t}$ represents for rotation error and translation error of the result \cite{xing2022reconstruction}. We further decompose $e_{t}$ into $e_{xy}$ and $e_{z}$, which are translation error in the xy-plane and along z-axis, to analyze the main cause of error.
        

\begin{table*}
        \centering

        \caption{\vspace{-0.5em} Accuracy Assessment of Closed-form Solution and Iterative Optimization Using Different Number of Data Sets}
        \vspace{-1.2em}
        \label{tab:calibration_results}
        \begin{tabular}{ccccccccc}
        \toprule
                \multirow{2}{*}{Number of data sets} & \multicolumn{2}{c}{$e_r$(\text{deg})} &  \multicolumn{2}{c}{$e_t$ (\text{mm})} & \multicolumn{2}{c}{$e_{xy}$ (\text{mm})}&  \multicolumn{2}{c}{$e_z$ (\text{mm})}\\ 
                \cmidrule{2-9}
                &Closed-form&Iterative&Closed-form&Iterative&Closed-form&Iterative&Closed-form&Iterative\\ \midrule
                4  & 9.02 & 0.64 & 2791 & 747 & 200 & 384  & 2783 & 641   \\ 
                5  & 0.62 & 0.48 & 27.43 & 26.22 & 9.11 & 8.56 & 25.87 & 24.78 \\ 
                6  & 0.47 & 0.39 & 16.96 & 16.52 & 6.06 & 5.74 & 15.83 & 15.49 \\ 
                8  & 0.35 & 0.31 & 9.66 & 9.46 & 3.21 & 2.89 & 9.11 & 8.79  \\ 
                10 & 0.30 & 0.27 & 7.57 & 7.54 & 2.73 & 2.67 & 7.06 & 7.05  \\ 
                15 & 0.21 & 0.20 & 5.41 & 5.38  & 1.89 & 1.86 & 5.07 & 5.05  \\ 
                20 & 0.18 & 0.17 & 4.71 & 4.67  & 1.65 & 1.63 & 4.41 & 4.38  \\ 
                30 & \textbf{0.14} & \textbf{0.13} & \textbf{3.38} & \textbf{3.35}  & \textbf{1.21} & \textbf{1.20} & \textbf{3.16} & \textbf{3.13}  \\ 
                \bottomrule
        \end{tabular}
        
        \begin{minipage}{0.95\linewidth}\vspace{0.2em}
        The standard deviations of rotation and translation errors over 50 trials of hand-eye calibration with different sample sizes. $e_r$ is the mean rotation error, $e_t$ is the mean translation error, $e_{xy}$ and $e_z$ are the mean translation errors in the xy-plane and along the z-axis, respectively.
        \end{minipage}
\vspace{-0.5em}
\end{table*}

        

\begin{table*}
        \centering
\vspace{-0.5em}
        \caption{\vspace{-0.5em} Comparison of Different Calibration Methods Using 3D Point Cloud}
        \vspace{-1.2em}
        \label{tab:accuracy_comparison}
        \begin{tabular}{ccccccccc}
        \toprule
        
                Method & $e_r$(\text{deg}) & $e_t$ (\text{mm}) & $e_{xy}$ (\text{mm})& $e_{z}$ (\text{mm}) & $E_{c}$(mm) & $E_{R}$(mm) & $e_t$/max($E_{c}$,$E_{R}$) & Runtime(s)\\ \midrule
                \textbf{PlaneHEC(ours)} & \textbf{0.13} & 3.35 & 1.20 & 3.13 & 10 & 0.05 & \textbf{0.34} & \textbf{0.28} \\
                LRBO\cite{li2024automatic} & 0.94 & 1.70 & 1.35 & 1.03 & 3 & 0.03 & 0.57 & {\textless}1 \\
                Murali, et al.\cite{murali2021situ} & 1$\sim$2 & \textbf{0.18} & \textbf{0.17} & \textbf{0.06} & 0.012 & 0.05 & 3.63 & NA \\
                Xing, et al.\cite{xing2022reconstruction} & 0.4 & 1.70 & NA & NA & 0.1 & 0.02 & 17 & 10.9 \\
                Peters, et al.\cite{peters2024robot} & 0.55 & 1.77 & NA & NA & 0.18 & 0.05 & 9.83 & 15984\\
                Zhong, et al.\cite{zhong2025hand} & 0.74 & 2 & NA & NA & 1 & 0.08 & 2 & NA \\
                
                \bottomrule
        \end{tabular}
        
        \begin{minipage}{0.95\linewidth}\vspace{0.2em}
        The standard deviations of rotation and translation errors of various methods. $e_r$ is the mean rotation error, $e_t$ is the mean translation error, $e_{xy}$ and $e_z$ are the mean translation errors in the xy-plane and along the z-axis, respectively. $E_{c}$ is camera's accuracy along z-axis and $E_{R}$ is robot arm's repeated positioning accuracy. The runtime does not include the data acquisition process.
        \end{minipage}
        \vspace{-1.5em}
\end{table*}

\subsection{Accuracy Assessment}

\subsubsection{Relative error analysis}
It is widely accepted that the ground truth of hand-eye relation is hardly available, so the calibration accuracy is usually evaluated in reconstruction and application. We kept the hand-eye transformation relationship between the robotic arm and the camera consistent, and observed the same plane from different perspectives to collect data. We collected 300 sets of point clouds and corresponding poses of the end-effector. 


Theoretically, $n$ transformation matrices should exhibit consistency, with their standard deviation quantifying calibration uncertainty~\cite{xing2022reconstruction}. Using 300 experimental datasets, we randomly sampled subsets (repeated 50 times) to compute transformation matrices. Table~\ref{tab:calibration_results} demonstrates decreasing standard deviations in both translation vectors and rotation matrices with increasing sample size. While PlaneHEC's theoretical minimum requires four datasets, practical implementation reveals significant translational errors at this threshold. Stable convergence to accurate hand-eye relationships occurs with $\geq$5 datasets, achieving 3.35~mm translational accuracy (surpassing camera's intrinsic precision) at 30 datasets. Error analysis shows dominant z-axis translational components compared to minimal x/y-axis deviations.

We compared the performance of our method against other baseline approaches. While the diverse platforms and test environments precluded a direct comparison of the relative translation errors, we instead evaluated the ratio of translation errors to the camera's relative accuracy. As shown in Table~\ref{tab:accuracy_comparison}, the results of dividing $e_{t}$ by the maximum of camera's accuracy $E_{c}$ and robot arm's repeated positioning accuracy $E_{R}$ indicated that our method outperformed the baseline methods in terms of both rotational and translation errors. PlaneHEC employs a plane fitting algorithm on a large set of points, effectively mitigating noise in the depth data. Consequently, the hand-eye calibration error (3.35mm) is lower than the inherent depth error of the camera (10mm). Moreover, the rotational error of the results is exceptionally low, consistently below 0.7 degrees, with optimal cases achieving a precision as fine as 0.13 degrees. Considering that the rotational error is minimal and that the translation error primarily manifests along the z-axis, it is practical to manually adjust the z-axis translation error in applied settings to attain better accuracy. Moreover, PlaneHEC only necessitated the robot's kinematic parameters and a relatively flat plane as a reference, showcasing substantial practical benefits.

\subsubsection{Plane Reconstruction}
Plane reconstruction serves as a method to assess the accuracy of hand-eye calibration by evaluating the alignment between the reconstructed plane and the real plane. Ideally, the reconstructed plane in the camera coordinate system should coincide precisely with the actual plane. 
In the static test, we fix the camera and obtain a point cloud to detect the plane. The normal vector of the plane and the distance from the plane to the center of the camera were calculated and the process was repeated 50 times to determine translation and rotation errors.
Subsequently, we adjust the robot arm's posture while maintaining the plane within the operational range of 0.3m to 0.8m, allowing the camera mounted on the robot arm to perform plane detection from various perspectives and distances to evaluate static error. 
The experimental results indicate that the average rotation error is approximately 0.27 degrees, while the average translation error is approximately 0.64mm.

\begin{figure}
        \centering
        \includegraphics[width=0.9\linewidth]{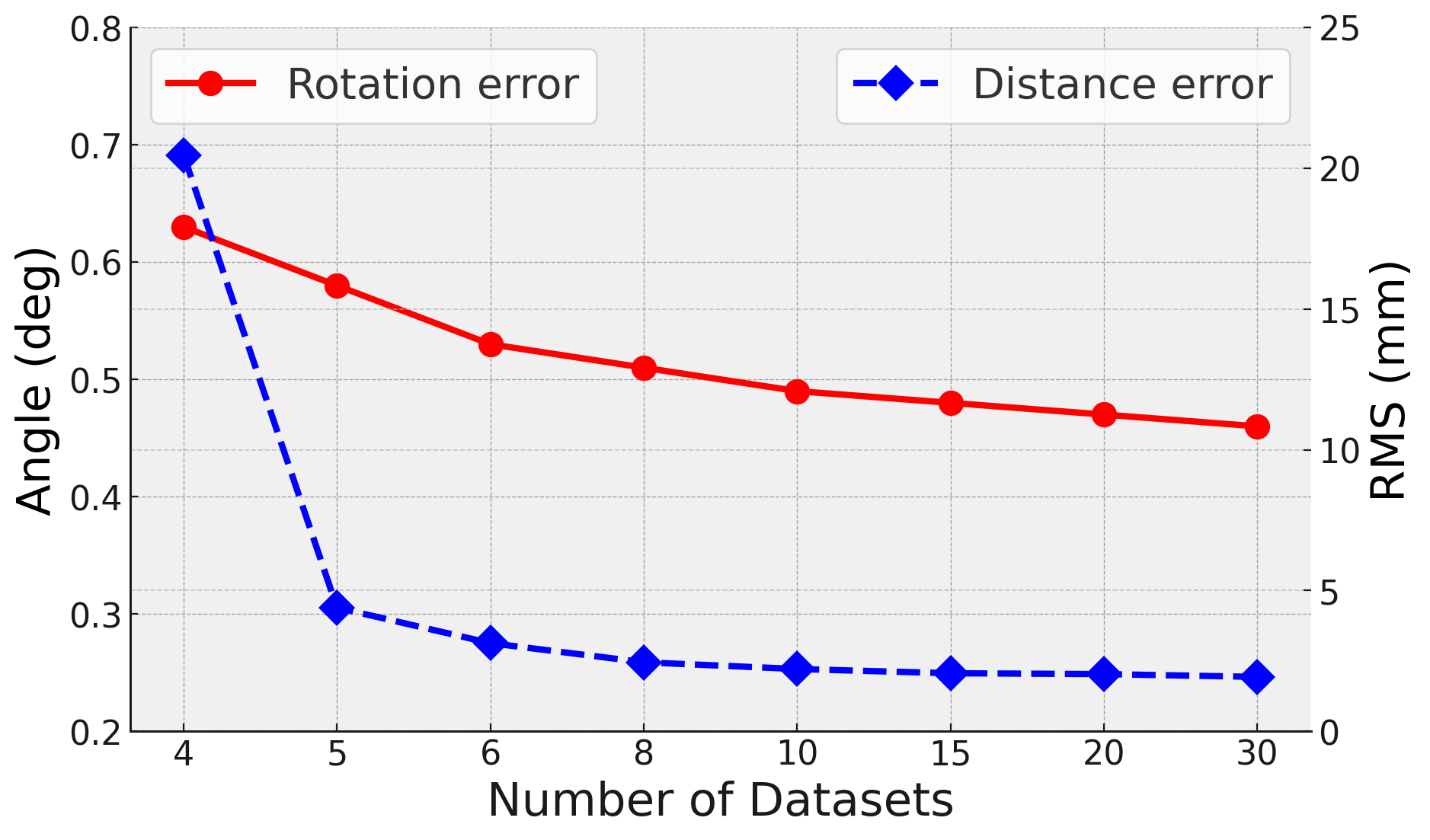}
        \vspace{-0.5em}
        \caption{\textbf{Dynamic error of plane reconstruction.} The rotation error decreases gradually as the number of data sets increases, while the distance error drops sharply.}
        \vspace{-1.5em}
        \label{fig:reconstruction_error}
\end{figure}

Subsequently, we conduct dynamic testing by transforming the planes observed from various perspectives into the base frame using the hand-eye calibration matrix and robot's pose, and evaluate the discrepancies between these calculated planes. We randomly selected some sets of data from a total of 300 sets to compute the hand-eye calibration matrix, using the remaining data for plane reconstruction. The results are shown in Figure~\ref{fig:reconstruction_error}. The rotation error gradually declines as the number of data sets increases and the distance error drops sharply when data sets increase from 4 to 5. After applying more than 10 data sets, both rotation error and distance error begin to converge and we have a minimum rotation error of 0.48 degrees and translation error of 2.04mm. The rotation error is comparable to the inherent static error of the plane detection algorithm, while the translation error is higher than the static error, yet remains within the camera's specified relative error tolerance of 10mm.

\subsubsection{Point Cloud and Robotic Arm Matching}
In addition to the aforementioned accuracy verification criteria, the accuracy of hand-eye calibration can also be evaluated by assessing the alignment between the point cloud and the robotic arm. When utilizing hand-eye calibration results in practical applications, if the robotic arm appears within the camera's field of view, observing the alignment between the point cloud and the physical model of the robotic arm serves as an effective verification method. In fact, several hand-eye calibration algorithms currently incorporate matching with the robotic arm model, using either depth~\cite{li2024automatic} or RGB~\cite{chen2023easyhec} data. Drawing inspiration from these approaches, we opted to use the robotic arm's physical model for validation.

In this experiment, we first imported the robotic arm's physical model into the ROS simulation environment and, based on the hand-eye calibration results, marked the camera's base coordinate system. We then obtained the point cloud using the pose of the robotic arm and the camera's depth data, transforming the point cloud into the base frame using the calibration results. By matching the point cloud with the simulated robotic arm model, we observed to what degree the point cloud and the robotic arm coincide.

\begin{figure}
        \centering
        \includegraphics[width=0.75\linewidth]{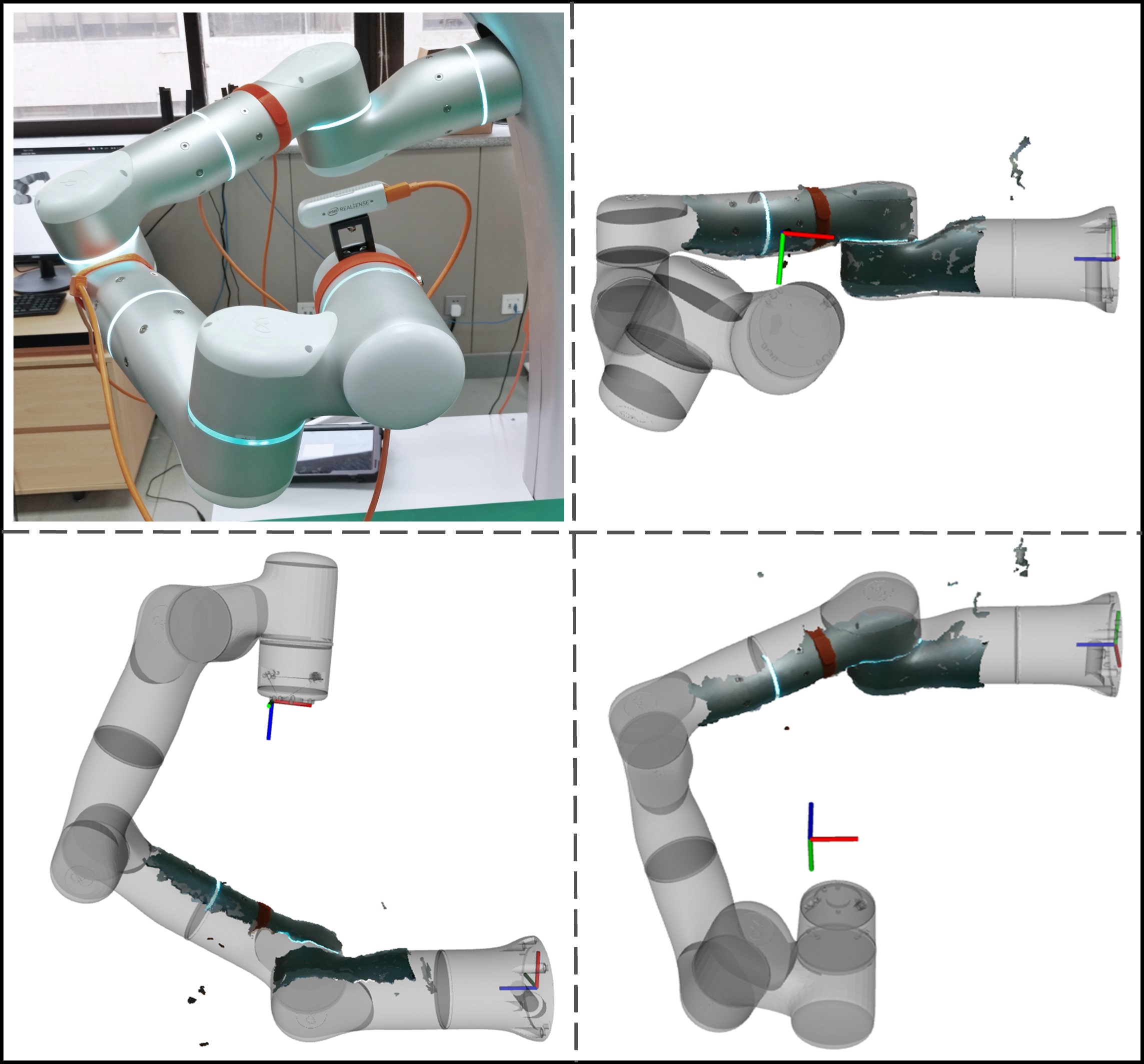}
        \vspace{-0.5em}
        \caption{\textbf{Point cloud alignment with the robotic arm model.} The top-left image presents an actual photograph of the robotic arm, while the remaining three images depict different viewpoints within the simulation environment. The point cloud data, acquired via an RGB-D camera, has been transformed utilizing the computed hand-eye calibration matrix.}
        \vspace{-0.5em}
        \label{fig:point_cloud_matching}
\end{figure}

We validated the hand-eye calibration result obtained from 30 sets of data. As depicted in Figure~\ref{fig:point_cloud_matching}, the results demonstrate that the point cloud's contour closely aligns with that of the robotic arm, and the LED strip is also tightly affixed to the joints of the robotic arm. We can conclude that our method achieves effective matching between the point cloud and the robotic arm.

\subsection{Ablation Study}
\subsubsection{Iterative vs. Closed-Form Error Comparison} We compared the relative errors between the closed-form and iterative solutions using the same dataset. The results are shown in Table~\ref{tab:calibration_results}, which demonstrated that the iterative approach yielded a smaller relative error compared to the closed-form solution. Generally speaking, the iterative optimization process has decreased $e_{R}$ and $e_{t}$ by 20\% and 8\% in average using different numbers of data sets, with a maximum of 93\% and 73\%. It is easy to conclude that the iteration has greatly improved the results, especially when only a few data sets are available. Besides, the improvement of rotation error is better than that in translation error, which can result from the inherent error of the camera.

\subsubsection{Analysis of Plane Detection}
\begin{figure}
        \centering
        \includegraphics[width=0.9\linewidth]{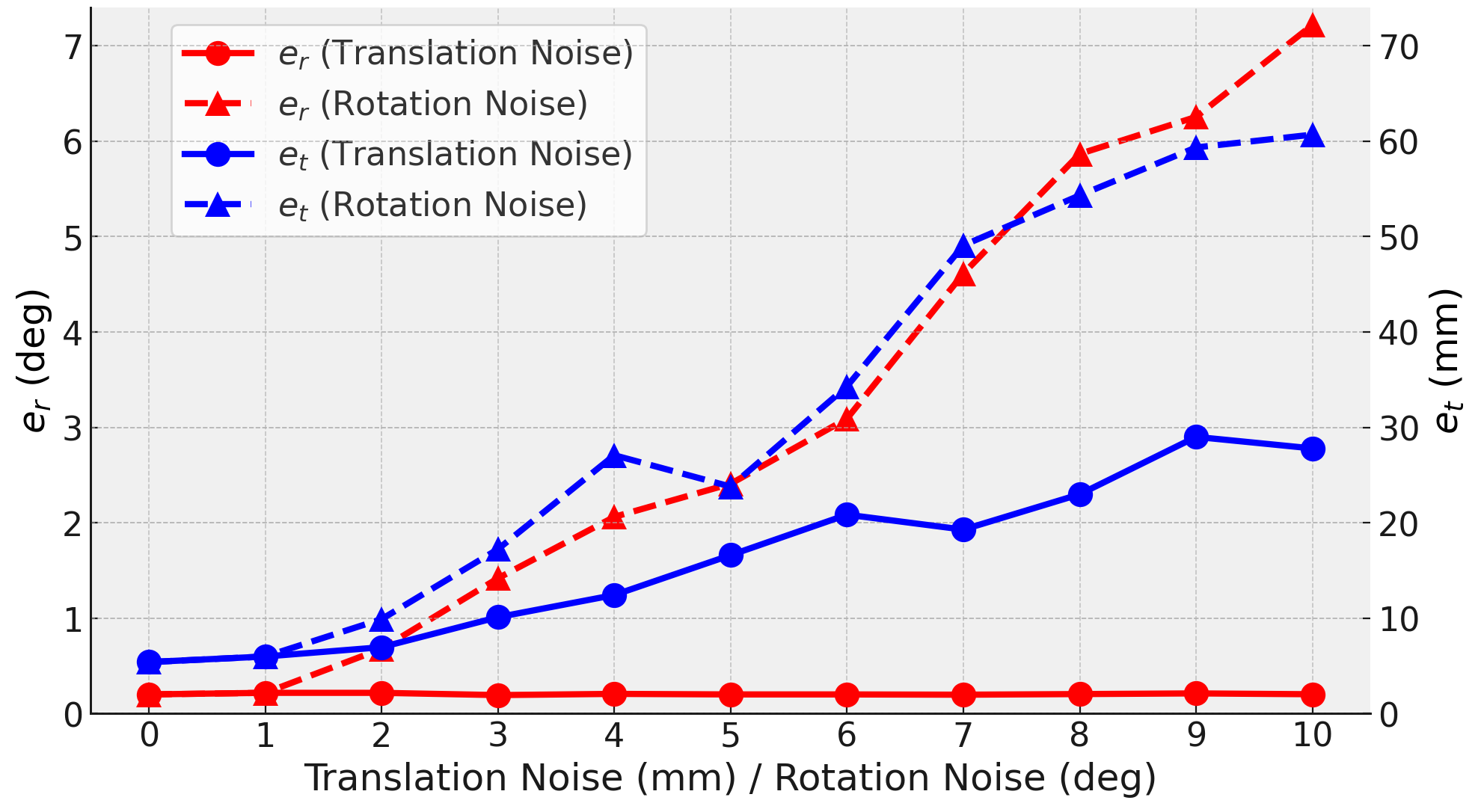}
        \vspace{-0.5em}
        \caption{\textbf{Impact of Plane Detection Error.} The translation noise in the plane parameters only affects the translation error, while the rotation noise affects both the rotation and translation errors.}
        \vspace{-1.5em}
        \label{fig:plane_error}
\end{figure}

In order to determine the possible effect of plane detection on the calibration results, we added Gaussian noise to the results of planar detection and compared the calibration results. Selecting 15 data sets as a batch, we repeated calibration 50 times to calculate the relative error. The results shown in Figure~\ref{fig:plane_error} indicate that the rotation error of the plane detection impacts both the rotational and translational accuracy, whereas the translational error only affects the translational accuracy.

\subsection{Runtime
}
As shown in Table~\ref{tab:accuracy_comparison}, the average processing time for 500 frames using MATLAB was approximately 0.01 seconds per frame, suggesting the arithmetic solving component of our approach is highly efficient and is the fastest method among other baseline methods. However, the majority of the runtime is attributed to the computationally intensive tasks of point cloud processing and plane detection. After optimization, the algorithm can be deployed for real-time online hand-eye calibration. Moreover, the algorithm exhibited rapid convergence, with an average of 8.6 iterations and a standard deviation not exceeding 1.5, demonstrating its robustness.


\section{Conclusions}



This paper presents PlaneHEC, a generalized hand-eye calibration method for multi-view robotic systems using depth cameras. The approach leverages plane-fitting algorithms to offer a fast and robust solution without the need for complex geometric models or external markers. Through simulations and real-world experiments, PlaneHEC performed better calibration accuracy compared to baseline methods, achieving translational errors as low as 3.35 mm, surpassing the camera's inherent depth error. Additionally, the method is highly sensitive to rotational transformations, maintaining rotational accuracy consistently below 0.7 degrees, making it ideal for applications requiring precise orientation adjustments. Comparative experiments showed that PlaneHEC algorithm performs best in terms of both accuracy and speed. Iterative optimization further refines the initial closed-form solution, enhancing overall performance. Future work will focus on improving real-time processing and adapting the method to multi-agent systems or embodied intelligence.





\bibliographystyle{IEEEtran}
\bibliography{references}

\end{document}